\relax
\documentclass[letterpaper]{article} % DO NOT CHANGE THIS
\usepackage{aaai22}  % DO NOT CHANGE THIS
\usepackage{times}  % DO NOT CHANGE THIS
\usepackage{helvet}  % DO NOT CHANGE THIS
\usepackage{courier}  % DO NOT CHANGE THIS
\usepackage[hyphens]{url}  % DO NOT CHANGE THIS
\usepackage{graphicx} % DO NOT CHANGE THIS
\usepackage{natbib}  % DO NOT CHANGE THIS AND DO NOT ADD ANY OPTIONS TO IT
\usepackage{caption} % DO NOT CHANGE THIS AND DO NOT ADD ANY OPTIONS TO IT
\DeclareCaptionStyle{ruled}{labelfont=normalfont,labelsep=colon,strut=off} % DO NOT CHANGE THIS
\usepackage{algorithm}
\usepackage{algorithmic}

\usepackage{newfloat}
\usepackage{listings}
\floatstyle{ruled}
\newfloat{listing}{tb}{lst}{}
\floatname{listing}{Listing}

\usepackage{amsfonts}
\usepackage{tabularx}
\usepackage{booktabs}
\usepackage{multirow}
\usepackage{bbding}
\usepackage{color}
\usepackage{xcolor}
\definecolor{g}{RGB}{78,172,91}
\definecolor{r}{RGB}{235,50,35}
\definecolor{citecolor}{HTML}{0071bc}
\usepackage{amsmath,bm}
\usepackage{xspace}

\makeatletter
\DeclareRobustCommand\onedot{\futurelet\@let@token\@onedot}
\def\@onedot{\ifx\@let@token.\else.\null\fi\xspace}

\def\eg{\emph{e.g}\onedot} 
\def\ie{\emph{i.e}\onedot}

\makeatother

\title{LCTR: On Awakening the Local Continuity of Transformer for \\ Weakly Supervised Object Localization}
\author {
    Zhiwei Chen\textsuperscript{\rm 1},
    Changan Wang\textsuperscript{\rm 2},
    Yabiao Wang\textsuperscript{\rm 2},
    Guannan Jiang\textsuperscript{\rm 3},\\
    Yunhang Shen\textsuperscript{\rm 2},
    Ying Tai\textsuperscript{\rm 2},
    Chengjie Wang\textsuperscript{\rm 2},
    Wei Zhang\textsuperscript{\rm 3},
    Liujuan Cao\textsuperscript{\rm 1}\thanks{Corresponding author.}
}
\affiliations {
    \textsuperscript{\rm 1}Media Analytics and Computing Lab, Department of Artificial Intelligence, School of Informatics, \\Xiamen University, China.
    \textsuperscript{\rm 2}Tencent Youtu Lab, Shanghai, China.
    \textsuperscript{\rm 3}CATL, China. \\
    zhiweichen.cn@gmail.com, \{changanwang, caseywang\}@tencent.com, jianggn@catl.com, \\
    \{odysseyshen, yingtai, jasoncjwang\}@tencent.com, zhangwei@catl.com, caoliujuan@xmu.edu.cn
}

\usepackage{bibentry}
\begin{document}
\maketitle

\begin{abstract}
Weakly supervised object localization (WSOL) aims to learn object localizer solely by using image-level labels.
The convolution neural network (CNN) based techniques often result in highlighting the most discriminative part of objects while ignoring the entire object extent.
Recently, the transformer architecture has been deployed to WSOL to capture the long-range feature dependencies with self-attention mechanism and multilayer perceptron structure.
Nevertheless, transformers lack the locality inductive bias inherent to CNNs and therefore may deteriorate local feature details in WSOL.
In this paper, we propose a novel framework built upon the transformer, termed LCTR (Local Continuity TRansformer), which targets at enhancing the local perception capability of global features among long-range feature dependencies.
To this end, we propose a relational patch-attention module (RPAM), which considers cross-patch information on a global basis.
We further design a cue digging module (CDM), which utilizes local features to guide the learning trend of the model for highlighting the weak local responses.
Finally, comprehensive experiments are carried out on two widely used datasets, \ie, CUB-200-2011 and ILSVRC, to verify the effectiveness of our method.

\end{abstract}

\begin{figure}[!t]
\centering
\includegraphics[width=1\linewidth]{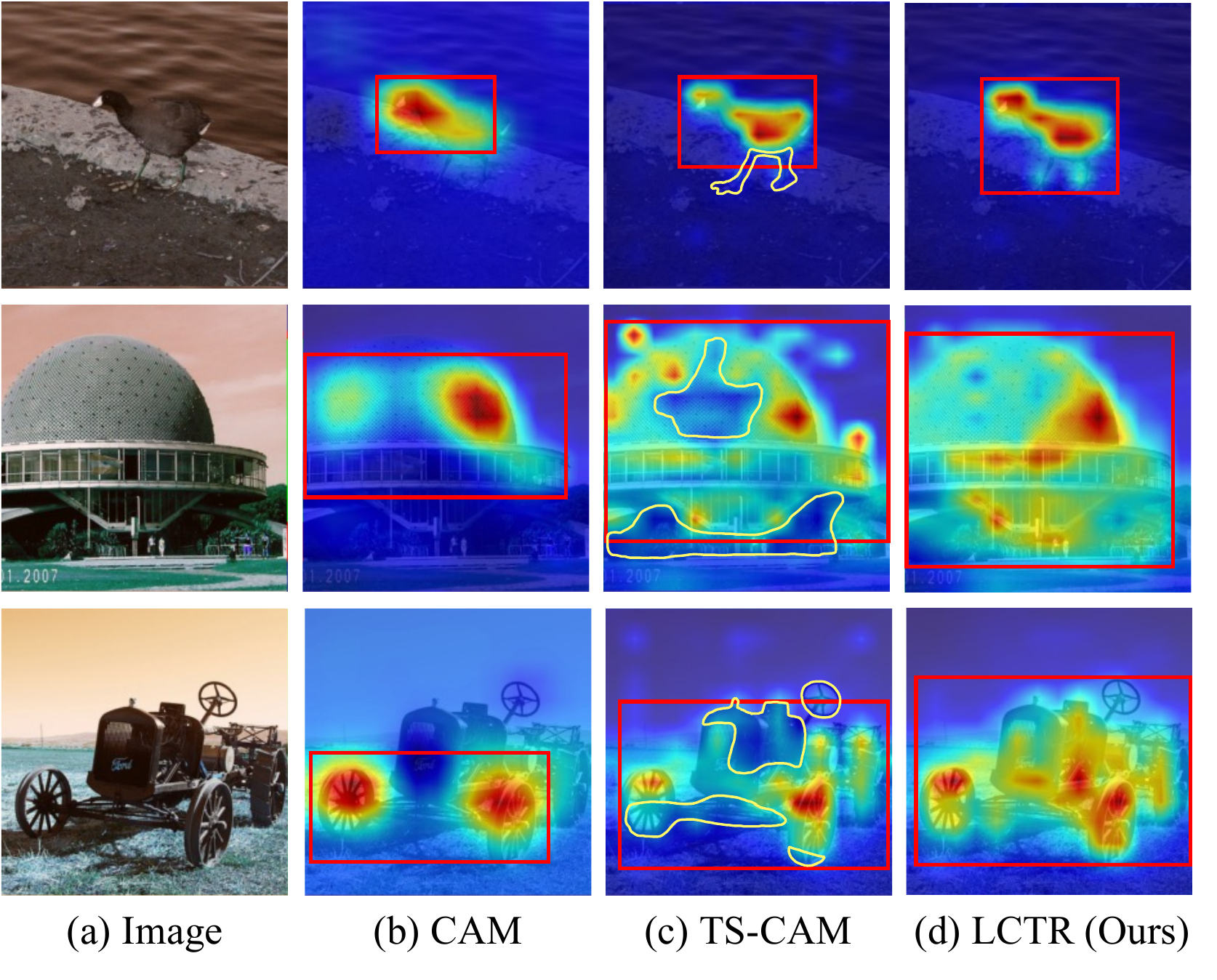}
\caption{
    Comparison of localization results on different methods:
    (a) Original images.
    (b) CNN-based method tends to be dominated by the most discriminative region.
    (c) Transformer-based method maintains coarse long-range dependencies while ignoring the local feature details (light yellow line).
    (d) The proposed LCTR not only considers finer local details but also retains global information.
    The predicted bounding boxes are in red. Best viewed in color.
}
\label{fig1}
\end{figure}

\section{Introduction}
Deep learning based methods have achieved unprecedented success in locating objects under a fully supervised setting~\cite{liu2016ssd,bochkovskiy2020yolov4,sun2021sparse,wang2021end}.
However, these methods rely on a large number of bounding box annotations, which are expensive to acquire.
Recently, the research on weakly supervised object localization (WSOL) has gained a significant momentum~\cite{Zhou_2016_CVPR,zhang2018adversarial,gao2021ts} since it can learn object localizers using only image-level labels.

%----------
The pioneering work \cite{Zhou_2016_CVPR} aggregated features from classification networks to generate class activation maps (CAM) for object localization.
Unfortunately, image classifiers tend to focus only on the most discriminative features to achieve high classification performance.
Therefore, the spatial distribution of feature responses may only cover the most discerning regions instead of the whole object range, which limits localization accuracy with large margins, as shown in Figure~\ref{fig1}(b).

%----------
To address this critical problem, many CAM-based approaches have been proposed, such as graph propagation~\cite{Zhu2017SPN}, data augmentation~\cite{kumar2017hide,yun2019cutmix}, adversarial erasing~\cite{zhang2018adversarial,choe2019attention,chen2021e2net} and spatial relation activation~\cite{xue2019danet,zhang2020inter,guo2021strengthen}.
However, those approaches do alleviate the partial activation issue, but in a compromised manner --- the essential philosophy behind it is first obtaining local features and then attempting to recover the non-salient regions to get full object extent.
In fact, the fundamental root of this issue is determined by the intrinsic nature of convolution neural networks (CNNs).
The CNN features with the local receptive field only capture small-range feature dependencies.
More recently, the transformer architecture~\cite{vaswani2017attention} has been developed in the field of computer vision~\cite{dosovitskiy2020image,wu2020visual,yuan2021tokens,touvron2021training,jiang2021transgan}, which shows that pure transformers can be as effective in feature extraction for image recognition as CNN-based architectures.
Notably, transformers with multi-head self-attention capture long-range dependencies, and retain more detailed information without downsampling operators, which naturally handles the partial activation problem in WSOL.
TS-CAM~\cite{gao2021ts} proposed token semantic coupled attention map from transformer structure, which captured long-range feature dependency among pixels for WSOL.
However, transformer-based methods lack the locality inductive bias inherent to CNNs, ignoring the local information, which leads to weak local feature response on the target object, as shown in Figure~\ref{fig1}(c).
Therefore, how to precisely mine local features in global representations for WSOL still remains an open problem.

%----------
In this paper, we propose a novel Local Continuity TRansformer (LCTR) for discovering entire objects of interest via end-to-end weakly supervised training.
The key idea of LCTR is to rearrange local-continuous visual patterns with global-connective self-attention maps, thereby bringing locality mechanism to transformer-based WSOL.
To this end, we first propose a relational patch-attention module (RPAM) to construct a powerful patch relation map, which takes advantage of the patch attention maps under the guidance of a global class-token attention map.
The RPAM maintains the cross-patch information and models a global representation with more local cues.
Second, a cue digging module (CDM) is designed succinctly to induce the model to highlight the weak local features (\eg, blurred object boundaries) by a hide-and-seek manner under a local view.
In the CDM, to reward the weak response parts, we propose to employ the erased strategy, and induce the learnable convolutional kernels to be weighted by the weak local features.
To validate the effectiveness of the proposed LCTR, we conduct a series of experiments on the challenging WSOL benchmarks.

%----------
Collectively, our main contributions are summarized as:
\begin{itemize}
	\item We propose a simple LCTR for WSOL, which greatly enhances the local perception capability of global self-attention maps among long-range feature dependencies.
	\item We design a relational patch-attention module (RPAM) by considering cross-patch information, which facilitates global representations.
	\item We introduce a cue digging module (CDM) that encodes weak local features by learnable kernels to highlight the local details of global representations.
	\item LCTR achieves new state-of-the-art performance on CUB-200-2011 and ILSVRC dataset with 79.2\% and 56.1\% Top-1 localization accuracy, respectively.
\end{itemize}

\begin{figure*}[!t]
\centering
\includegraphics[width=1\linewidth]{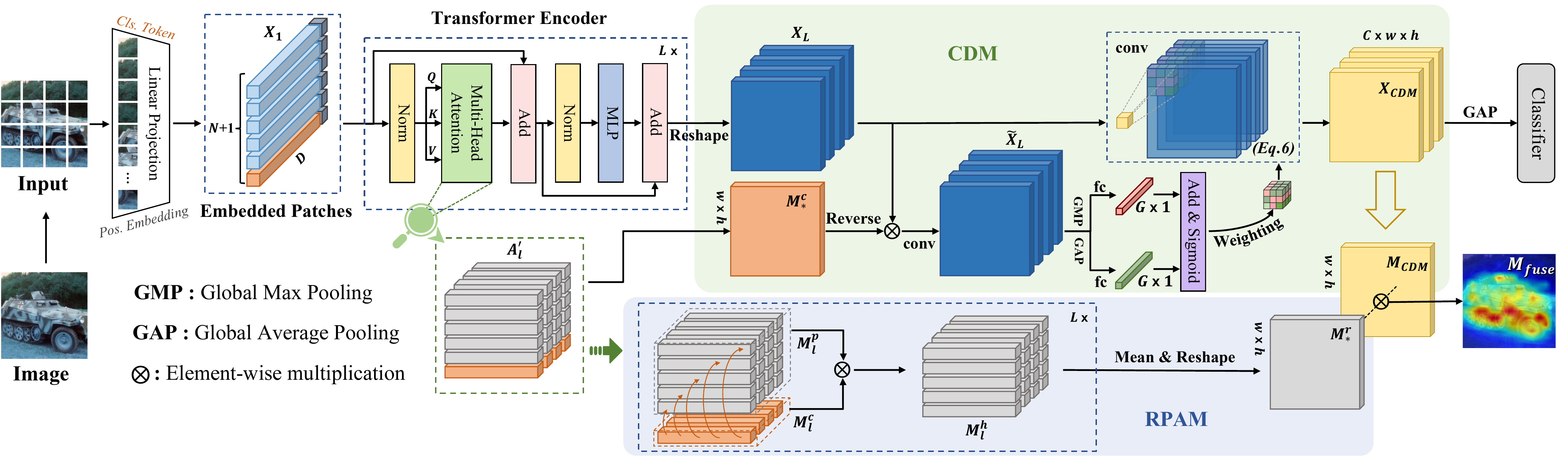}
\caption{
    Overview of the proposed LCTR, which consists of vision transformer backbone for feature extraction, relational patch-attention module (RPAM) and cue digging module (CDM).
}
\label{fig2}
\end{figure*}

\section{Related Work}
%----------
\textbf{CNN-based Methods for WSOL}.
WSOL aims to learn object localizers with solely image-level supervision.
There are many state-of-the-art methods based on the CNN structure.
A representative pipeline of CNN-based WSOL is to aggregate deep feature maps with a class-specific fully connected layer to produce class attention maps (CAMs), from which final predicted bounding boxes are extracted~\cite{Zhou_2016_CVPR}.
Later on, the last fully connected layer is dropped for simplifying~\cite{hwang2016self}.
Unfortunately, CAMs tend to be dominated by the most discriminative object part.
Therefore, different extensions~\cite{selvaraju2017grad,chattopadhay2018grad,xue2019danet,zhang2020inter} have been proposed to improve the generation process of localization maps in order to recover the non-salient regions.
HaS~\cite{kumar2017hide} and CutMix~\cite{yun2019cutmix} adopted a random-erasing strategy from input images to force the classification networks to focus on relevant parts of objects.
ACoL~\cite{zhang2018adversarial} introduced two adversarial classification classifiers to locate different object parts and discovered the complementary regions belonging to the same objects or categories.
ADL~\cite{choe2019attention} further promoted the localization maps by applying dropout on multiple intermediate feature maps.
Besides the erasing strategy, DANet~\cite{xue2019danet} used a divergent activation method to learn better localization maps.
SPG~\cite{zhang2018self} and I$^2$C~\cite{zhang2020inter} introduced the constraint of pixel-level correlations into the WSOL network.
SPA~\cite{pan2021unveiling} leveraged structure information incorporated in convolutional features for WSOL.
Some other methods (\eg, GC-Net~\cite{Lu2020GeometryCW}, PSOL~\cite{zhang2020rethinking}, SPOL~\cite{wei2021shallow} and SLT-Net~\cite{guo2021strengthen}) divided WSOL into two independent sub-tasks, including classification and the class-agnostic localization.

%----------
These studies alleviate the problem by extending from local activations to global ones in an implicit way, which is difficult to balance the image classification and the object localization.
In fact, CNNs are prone to capture partial semantic features with local receptive fields, which belongs to the principal problem of CNNs.
The problem of how to explore global cues from local receptive fields still exists.
In this paper, we introduce a transformer-based structure, where the local-continuity and long-range feature dependencies can be simultaneously activated.

%----------
\textbf{Transformer-based Methods for WSOL}.
The transformer model~\cite{vaswani2017attention} is proposed to handle sequential data in the field of natural language processing.
Recent studies also reveal its effectiveness for computer vision tasks~\cite{dosovitskiy2020image,beal2020toward,carion2020end,zheng2021rethinking,hu2021istr}.
Since the local information extracted by the CNNs is deficient, various methods adopt the self-attention mechanism to capture the long-range feature dependencies.
ViT~\cite{dosovitskiy2020image} applied the pure transformer directly to sequences of image patches for exploring spatial correlation on the image classification task.
DETR~\cite{carion2020end} employed a transformer encoder-decoder architecture for the object detection task.
As a pioneered work in WSOL, TS-CAM~\cite{gao2021ts} proposed a semantic coupling strategy based on Deit~\cite{touvron2021training} to fuse the patch tokens with the semantic-agnostic attention map to achieve semantic-aware localization results, which has inspired many scholars to study transformer in weakly supervised object localization.

%----------
Despite of the progress, TS-CAM is relatively rough as it single-mindedly tries to get the long-range features but ignores the local information.
Compared with the existing methods, our LCTR retains long-dependent features while mining for detailed feature cues based on the transformer structure for WSOL.

\section{Methodology}
We first present the overview of the proposed LCTR, then give a detailed description of RPAM and CDM, and finally incorporate them with the transformer structure in a joint optimization framework, as shown in Figure~\ref{fig2}.

\subsection{Overview}
In accordance with the long-range info-preserving ability of the transformer architecture, LCTR is designed to offer precise localization maps for WSOL.
We denote the input images as $I=\{(I_i,y_i)\}_{i=0}^{M-1}$, where $y_i\in \{0,1,\dots ,C-1\}$ indicates the label of the image $I_i$, $M$ and $C$ are the number of images and classes, respectively.
%
% an image of size $H \times W$, where $H$, $W$ denote its height and width.
%
%For simplicity, we omit the mini-batch dimension in this notation.
%
We first split $I_i$ into $N$ same-sized patches $x_p\in \mathbb{R}^{1\times D}$, where $D$ denotes the dimension of each patch.
We set $N=w \times h$, $w=W/P$ and $ h=H/P$, where $P$ is the width/height of a patch, $H$ and $W$ denote image height and width.
For simplicity, we omit the mini-batch dimension.
A learnable class token $x_{cls}\in \mathbb{R}^{1\times D}$ is embedded into the patches.
These patches are flattened and linearly projected before being fed to $L$ sequential transformer blocks, which can be formulated as:
%__________
\begin{equation}
	X_1=[x_{cls};\mathcal{F}(x_p^1);\mathcal{F}(x_p^2);\cdots ;\mathcal{F}(x_p^N)]+\mathcal{P},
\end{equation}
%__________
where $X_1$ denotes the input of the first transformer block, $\mathcal{P} \in \mathbb{R}^{(N+1)\times D}$ is the position embedding and $\mathcal{F}$ is a linear projection.
In particular, the proposed RPAM is employed in each transformer block to obtain a patch relation map $M^r_*$, which aggregates cross-patch information on a global basis.

%----------
Denote ${X_L} \in \mathbb{R}^{N\times D}$ as the output feature of the last transformer block.
We reshape ${X_L} \in \mathbb{R}^{D\times w \times h}$ and apply the proposed CDM for further highlighting weak local responses.
After that we obtain the feature map $\mathrm{X}_{CDM} \in \mathbb{R}^{C\times w \times h}$.
Finally, the $\mathrm{X}_{CDM}$ are fed to a global average pooling (GAP) layer~\cite{lin2013network} followed by a softmax layer to predict the classification probability $p \in \mathbb{R}^{1\times C}$.
The loss function is defined as
%__________
\begin{equation}
	\mathcal{L}=-\log p.
	\label{loss}
\end{equation}
%__________

During testing, we extract the object map ${M_{CDM}} \in \mathbb{R}^{w \times h}$ from $\mathrm{X}_{CDM}$ according to the predicted class and obtain the final localization map by element-wise multiplication, given as
%__________
\begin{equation}
	M^{fuse}=M_{CDM} \otimes M^r_*.
\end{equation}
%__________
The $M^{fuse}$ is then resized to the same size as the original images by linear interpolation.
For a fair comparison, we apply the same strategy detailed in CAM~\cite{Zhou_2016_CVPR} to produce the object bounding boxes.

\subsection{Relational Patch-Attention Module}
\begin{figure}[!t]
\centering
\includegraphics[width=1\linewidth]{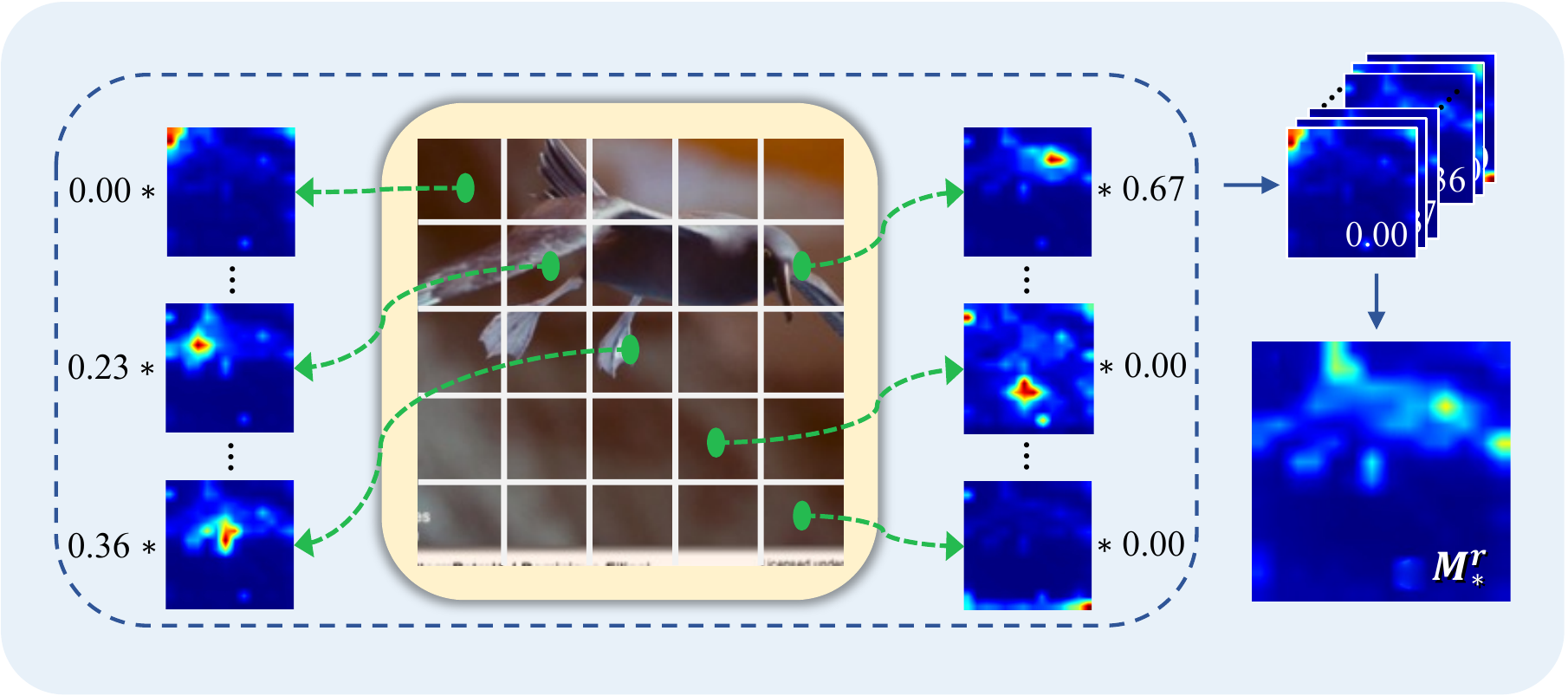}
\caption{
     The RPAM aggregates all patch attention maps based on the scores (values) of the class-token attention map to learn local visual patterns.
}
\label{fig3}
\end{figure}

The proposed relational patch-attention module (RPAM) (Figure~\ref{fig3}) strengthens the global feature representation from two stages:
First, we utilize the attention vectors of the class token in the transformer block to generate a global class-token attention map.
To fully exploit the feature dependencies of the transformer structure, we then use all the attention vectors of the patches containing the correlation between local features to generate a patch relation map under the guidance of the class-token attention map.

%----------
In the $l$-th transformer block, we hypothesize that the output feature map is ${X_l} \in \mathbb{R}^{(N+1)\times D}$.
The attention matrix ${A_l}\in \mathbb{R}^{S\times (N+1) \times (N+1)}$ of multi-head self-attention module in the block is formulated as:
%__________
\begin{equation}
	{A_l}=\mathrm{Softmax}\left ( \frac{\bm{Q}_l\cdot \bm{K^{\top}}_l}{\sqrt{D/S}}  \right ),
\end{equation}
%__________
where $\bm{Q}_l$ and $\bm{K}_l$ denote the queries and keys projected by ${X_l}$ of self-attention operation in $(l-1)$-th transformer block, respectively.
$S$ represents the number of head and $\top$ is a transpose operator.

%----------
At this point, we first take the average operator to ${A}_l$ based on $S$ heads to obtain ${A^\prime_l} \in \mathbb{R}^{(N+1) \times (N+1)}$.
Then, the class-token attention vector $M^c_l \in \mathbb{R}^{1 \times (N+1)}$ is extracted from ${A^\prime_l}$.
The $M^c_l$ reveals how much each patch contributes to the object regions for image classification.
Unfortunately, this map simply captures the global interactions of the class token to all patches, while ignoring the cross-patch correlations, which affects the modeling of local features.
To remedy it, we take advantage of the patch attention map $M^p_l\in \mathbb{R}^{(N+1) \times N}$ in ${A^\prime_l}$ to structure a patch relation vector $M^r_l$ under the guidance of $M^c_l$.
The $M^p_l$ learns the correlation between each patch but couldn't tell which one is more important.
Therefore, we weight each patch attention map by multiplying $M^p_l$ by $M^c_l$ to obtain a new map $M^h_l\in \mathbb{R}^{(N+1) \times N}$.
Note that $M^c_l$ is reshaped ($\mathbb{R}^{(N+1) \times 1}$) before the multiplication.
After that, we squeeze the first dimension of $M^h_l$ to a vector ($M^r_l \in \mathbb{R}^{1 \times N}$) by an average operation.
The final patch relation map $M^r_\ast$ is calculated by
%__________
\begin{equation}
	M^r_\ast=\Gamma^{w\times h} (\frac{1}{L} \sum_l M_{l}^r),
\end{equation}
%__________
where $\Gamma^{w\times h}(\cdot)$ indicates the reshape operator which coverts the vector ($\mathbb{R}^{1 \times N}$) to the map ($\mathbb{R}^{w \times h}$).

%----------
The patch relation map $M^r_\ast$ obtains the long-range dependencies that depends on the class-token attention vector $M^c_l$.
Aggregating cross-patch information from the patch attention maps, $M^r_\ast$ facilitates better global representations of the object without extra parameters in a simple way.

\subsection{Cue Digging Module}
RPAM considers the cross-patch information by using the class-token attention map from self-attention mechanism block of the transformer structure, but it is vulnerable if the transformer gets a poor class-token attention map.
We thus further propose a cue digging module (CDM) to supply the long-range features based on a hide-and-seek manner.

%----------
Inspired by erasing-based methods that remove the most discriminative parts of the target object to induce the model to cover the integral extent of the object, we erase the object regions based on the global class-token attention map, leaving the weak response ones and the background.
Then by weighting the learnable convolution kernels in the CDM on the basis of them, we shift part of the attention to object regions with weak responses.
With the help of weighted kernels, we can highlight the local details as a supplement to the global representations.

%----------
Specifically, we convert the patch parts of class-token attention vectors to the map $\tilde{M}^c_* \in \mathbb{R}^{w \times h}$, and apply it to the feature map ${X_L}$ by spatial-wise multiplication after being reversed.
Note that $\tilde{M}^c_*$ is calculated by $\tilde{M}^c_*=\frac{1}{L}  {\textstyle \sum_{l}} M^c_l $.
The feature map then passes through a convolutional layer to generate a new feature map ${\tilde{X}_L} \in \mathbb{R}^{D \times w \times h}$.
Next, we score the features into $G$ parts corresponding to $G$ learnable convolution kernels.
In particular, we apply two separate operators, the global average pooling and max pooling, to ${\tilde{X}_L}$.
Then the feature maps are vectorized and sent to a fully connected layer, respectively.
Besides, we add them together and apply a sigmoid function.
In this light, we obtain the scores as $\{S_g\mid g=1,2,\dots ,G\}$.
Finally, ${X_L}$ passes through a convolutional layer with weighted convolution kernels by the scores $S_g$ for encouraging the model to learn the object regions with weak responses, which can be formulated as
%__________
\begin{equation}
	{X_{SGR}}={X_L} \sum_{g=1}^{G} S_g W^e_g,
\end{equation}
%__________
where $W^e_g \in \mathbb{R}^{D \times C \times k_{w} \times k_{h}}$ denotes the kernel weights of the convolutional layer, which are initialized with kaiming uniform initialization~\cite{he2015delving}.
$k_{w}$, $k_{h}$ represent the width and height of the kernels, respectively.

%----------
The convolution layer with the weighted kernels is applied to the global feature map $\mathrm{X}_L$ drawn from the transformer structure.
Once the loss $\mathcal{L}$ in Eq.~\ref{loss} is optimized, the weighted convolution kernels become more sensitive to the features (\ie, the weak response features of object regions) that favor image classification.
In this way, the model pays more attention to local cues and forms better global representations for the target object.

\section{Experiments}
\subsection{Experimental Settings}
\textbf{Datasets.}
We evaluate the proposed methods on two challenging datasets, including CUB-200-2011~\cite{wah2011caltech} and ILSVRC~\cite{russakovsky2015imagenet}.
We only use image-level labels for training.
CUB-200-2011 is a fine-grained bird dataset of $200$ categories, which contains $5,994$ images for training and $5,794$ for testing.
ILSVRC has about $1.2$ million images in the training set and $50,000$ images in the validation set, with a total of $1,000$ different categories.

%----------
\textbf{Evaluation Metrics.}
Following previous methods~\cite{Zhou_2016_CVPR,russakovsky2015imagenet}, we adopt the Top-1/Top-5 classification accuracy (Top-1/Top-5 \textit{Cls.}), Top-1/Top-5 localization accuracy (Top-1/Top-5 \textit{Loc.}) and localization accuracy with known ground-truth class (\textit{Gt-k.}) as our evaluation metrics.
Specifically, Top-1/Top-5 \textit{Cls.} is correct if the Top-1/Top-5 predicted category contains the correct label.
\textit{Gt-k.} is correct when the intersection over union (IoU) between the ground-truth and the prediction is larger than $0.5$, and does not consider whether the predicted category is correct.
Top-1/Top-5 \textit{Loc.} is correct when Top-1/Top-5 \textit{Cls.} and \textit{Gt-k.} are both correct.

%%-----------Localization on CUB-200-2011 validation set.
\begin{table}[!t]
  \centering
    \begin{tabularx}{\linewidth}{l|c|c|c|X<{\centering}}
		\toprule
		\multirow{2}{*}{Methods (Yr)} & \multirow{2}{*}{Backbone} & \multicolumn{3}{c}{Loc. Acc} \\ \cline{3-5} 
	     &  &Top-1&Top-5&Gt-k.\\
	    \hline
	    CAM \small('16) & GoogLeNet & 41.1 & 50.7 & 55.1 \\
	    SPG \small('18) & GoogLeNet & 46.7 & 57.2 & - \\
	    RCAM \small('20) & GoogLeNet & 53.0 & - & 70.0 \\
	    DANet \small('19) & InceptionV3 & 49.5 & 60.5 & 67.0 \\
	    ADL \small('19) & InceptionV3 & 53.0 & - & - \\
	    PSOL \small('20) & InceptionV3 & 65.5 & - & - \\
	    SPA \small('21) & InceptionV3 & 53.6 & 66.5 & 72.1 \\
		SLT-Net \small('21) & InceptionV3 & 66.1 & - & 86.5 \\

		\hline
	    CAM \small('16) & VGG16 & 44.2 & 52.2 & 56.0 \\
	    ADL \small('19) & VGG16 & 52.4 & - & 75.4 \\
	    ACoL \small('18) & VGG16 & 45.9 & 56.5 & 59.3 \\
	    SPG \small('18) & VGG16 & 48.9 & 57.2 & 58.9 \\
	    DANet \small('19) & VGG16 & 52.5 & 62.0 & 67.7 \\
	    MEIL \small('20) & VGG16 & 57.5 & - & 73.8 \\
	    PSOL \small('20) & VGG16 & 66.3 & - & - \\
	    RCAM \small('20) & VGG16 & 59.0 & - & 76.3 \\
	    GC-Net \small('20) & VGG16 & 63.2 & - & - \\
	    SPA \small('21) & VGG16 & 60.2 & 72.5 & 77.2 \\
	    SLT-Net \small('21) & VGG16 & 67.8 & - & 87.6 \\

	    \hline
	    TS-CAM \small('21)& Deit-S & \textcolor{g}{\textbf{71.3}}  &  \textcolor{g}{\textbf{83.8}} & \textcolor{g}{\textbf{87.7}} \\
	    LCTR (Ours)  & Deit-S & \textcolor{r}{\textbf{79.2}} & \textcolor{r}{\textbf{89.9}} & \textcolor{r}{\textbf{92.4}} \\  	 	 
	    \bottomrule
    \end{tabularx}
 	\caption{Localization accuracy on the CUB-200-2011 test set.  The results of the first and second are shown in \textcolor{r}{red} and \textcolor{g}{green}, respectively.}
  \label{tab1}
\end{table}

%%-----------Localization on ILSVRC validation set.
\begin{table}[!t]
  \centering
    \begin{tabularx}{\linewidth}{l|c|c|c|X<{\centering}}
		\toprule
		\multirow{2}{*}{Methods (Yr)} & \multirow{2}{*}{Backbone} & \multicolumn{3}{c}{Loc. Acc} \\ \cline{3-5} 
	     &  &Top-1&Top-5&Gt-k.\\
	    \hline
	    CAM \small('16) & VGG16 & 38.9 & 48.5 & - \\
	    ACoL \small('18) & VGG16 & 45.8 & 59.4 & 63.0 \\
	    CutMix \small('19) & VGG16 & 42.8 & 54.9 & 59.0 \\
	    ADL \small('19) & VGG16 & 44.9 & - & - \\
	    I$^2$C \small('20) & VGG16 & 47.4 & 58.5 & 63.9 \\
	    MEIL \small('20) & VGG16 & 46.8 & - & - \\
	    RCAM \small('20) & VGG16 & 44.6 & - & 60.7 \\
	    PSOL \small('20) & VGG16 & 50.9 & 60.9 & 64.0 \\
	    SPA \small('21) & VGG16 & 49.6 & 61.3 & 65.1 \\
	    SLT-Net \small('21) & VGG16 & 51.2 & 62.4 & 67.2 \\
		\hline
	    CAM \small('16) & InceptionV3 & 46.3 & 58.2 & 62.7 \\
	    SPG \small('18) & InceptionV3 & 48.6 & 60.0 & 64.7 \\
	    ADL \small('19) & InceptionV3 & 48.7 & - & - \\
	    ACoL \small('18) & GoogLeNet & 46.7 & 57.4 & - \\
	    DANet \small('19) & GoogLeNet & 47.5 & 58.3 & - \\
	    RCAM \small('20) & GoogLeNet & 50.6 & - & 64.4 \\
	    MEIL \small('20) & InceptionV3 & 49.5 & - & - \\
	    I$^2$C \small('20) & InceptionV3 & 53.1 & 64.1 & \textcolor{g}{\textbf{68.5}} \\
	    GC-Net \small('20) & InceptionV3 & 49.1 & 58.1 & - \\
	    PSOL \small('20) & InceptionV3 & 54.8 & 63.3 & 65.2 \\
	    SPA \small('21) & InceptionV3 & 52.8 & 64.3 & 68.4 \\
	    SLT-Net \small('21) & InceptionV3 & \textcolor{g}{\textbf{55.7}} & \textcolor{g}{\textbf{65.4}} & 67.6 \\
	    \hline
	    TS-CAM \small('21)& Deit-S & 53.4  &  64.3 & 67.6 \\
	    LCTR (Ours) & Deit-S & \textcolor{r}{\textbf{56.1}} & \textcolor{r}{\textbf{65.8}} & \textcolor{r}{\textbf{68.7}} \\
	    \bottomrule
    \end{tabularx}
 	\caption{Localization accuracy on the ILSVRC validation set. The results of the first and second are shown in \textcolor{r}{red} and \textcolor{g}{green}, respectively.}
  \label{tab_2}
\end{table}

%%-----------Classification on CUB validation set.
\begin{table}[!t]
  \centering
    \begin{tabularx}{\linewidth}{l|c|X<{\centering}|X<{\centering}}
		\toprule
		\multirow{2}{*}{Methods (Yr)} & \multirow{2}{*}{Backbone} & \multicolumn{2}{c}{Cls. Acc} \\ \cline{3-4} 
	     &  &Top-1&Top-5\\
	     
	    \hline
	    CAM \small('16) & GoogLeNet & 73.8 & 91.5 \\
	    RCAM \small('20) & GoogLeNet & 73.7 & - \\
	    DANet \small('19) & InceptionV3 & 71.2 & 90.6 \\
	    ADL \small('19) & InceptionV3 & 74.6 & - \\
	    SLT-Net \small('21) & InceptionV3 & 76.4 & - \\
	    
	    \hline
	    CAM \small('16) & VGG16 & 76.6 & 92.5 \\
	    ACoL \small('18) & VGG16 & 71.9 & - \\
	    ADL \small('19) & VGG16 & 65.3 & - \\
	    DANet \small('19) & VGG16 & 75.4 & 92.3 \\
	    SPG \small('18) & VGG16 & 75.5 & 92.1 \\
	    MEIL \small('20) & VGG16 & 74.8 & - \\
	    RCAM \small('20) & VGG16 & 75.0 & - \\
	    SLT-Net \small('21) & VGG16 & 76.6 & - \\
		\hline
	    
	    \hline
	    TS-CAM \small('21)& Deit-S & \textcolor{g}{\textbf{80.3}} & \textcolor{g}{\textbf{94.8}} \\
	    LCTR (Ours) & Deit-S & \textcolor{r}{\textbf{85.0}} & \textcolor{r}{\textbf{97.1}} \\  	 
	    \bottomrule
    \end{tabularx}
 	\caption{Classification accuracy on the CUB-200-2011 test set. The results of the first and second are shown in \textcolor{r}{red} and \textcolor{g}{green}, respectively.}
  \label{tab3}
\end{table}

%----------
\textbf{Implementation Details.}
We adopt the Deit~\cite{touvron2021training} as the backbone network, which is pre-trained on ILSVRC~\cite{russakovsky2015imagenet}.
Particularly, we replace the MLP head with our proposed CDM.
Finally, a GAP layer and a softmax layer are added on the top of the convolutional layers.
The input images are randomly cropped to $224 \times 224$ pixels after being resized to $256 \times 256$ pixels.
We adopt AdamW~\cite{loshchilov2017decoupled} with $\epsilon$=$1e$-$8$, $\beta_1$=$0.9$, $\beta_2$=$0.99$ and weight decay of $5e$-$4$.
On CUB-200-2011, we use a batch size of $128$ with a learning rate of $5e$-$5$ to train the model for 80 epochs.
For ILSVRC, the training process lasts $14$ epochs with a batch size of $256$ and a learning rate of $5e$-$4$.
After meticulous experiments, we set $G$=4 in the CDM.
All the experiments are performed with four Nvidia Tesla V100 GPUs using the PyTorch toolbox.

\begin{figure*}[!t]
\centering
\includegraphics[width=1\linewidth]{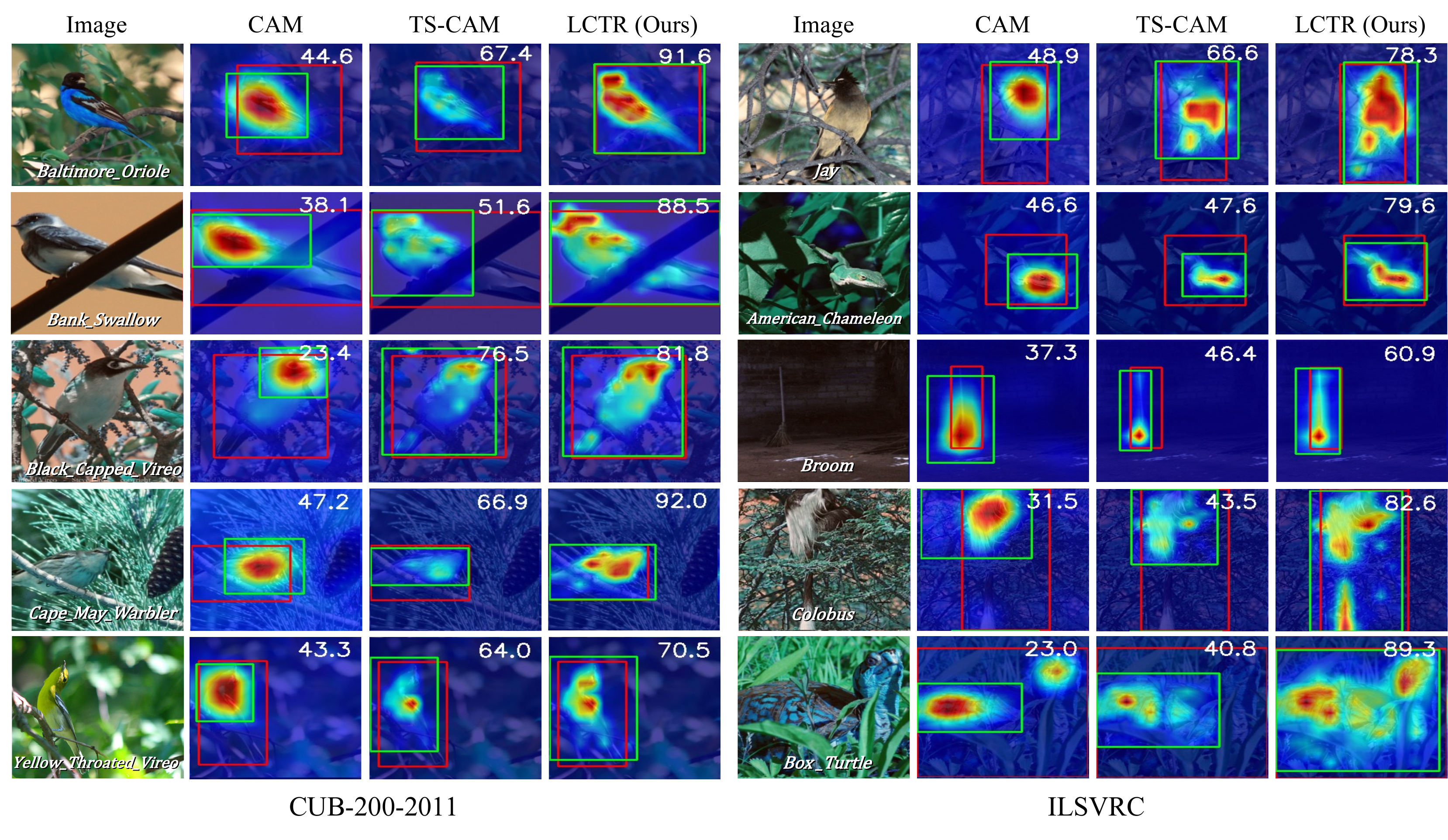}
\caption{
    Visual comparisons of localization results on different methods.
    \textbf{1st Column}: Input images.
    \textbf{2nd Column}: Results of CAM based on CNNs.
    \textbf{3rd Column}: Results of TS-CAM based on Transformer.
    \textbf{4th Column}: Results of our LCTR.
    Note that the groundtruth bounding boxes are in \textcolor{red}{red}, the predictions are in \textcolor{green}{green}, and the IoU values (\%) are shown in white text.
}
\label{fig4}
\end{figure*}

\subsection{Comparison with the State-of-the-Arts}
%----------
\textbf{Localization.}
We first compare the proposed LCTR with the SOTAs on the localization accuracy on the CUB-200-2011 test set, as illustrated in Table~\ref{tab1}.
We observe that LCTR outperforms the baseline (\ie, TS-CAM~\cite{gao2021ts}) by 7.9\% in terms of Top-1 \textit{Loc.}, and is obviously superior to these CNN-based methods.
Besides, Table~\ref{tab_2} illustrates the localization accuracy on the ILSVRC validation set.
It reports 0.4\% performance improvement over the state-of-the-art SLT-Net~\cite{guo2021strengthen}.

%----------
\textbf{Classification.}
Table~\ref{tab3} and Table~\ref{tab3} show the Top-1 and Top-5 classification accuracy on the CUB-200-2011 test set and ILSVRC validation set, respectively.
For the fine-grained recognition dataset CUB-200-2011, LCTR achieves remarkable performance of 85.0\%/97.1\% on Top1/Top-5 \textit{Acc.}.
In addition, LCTR obtains comparable results with SLT-Net~\cite{guo2021strengthen} on Top-1 \textit{Acc.} and surpasses other methods significantly on the ILSVRC validation set.
Note that SLT-Net used a separated localization-classification framework, it cannot retain the global information for the objects in the individual classification network.
To sum up, the proposed LCTR can greatly improve the quality of object localization while keeping high classification performance.

%%-----------Classification on ILSVRC validation set.
\begin{table}[!t]
  \centering
    \begin{tabularx}{\linewidth}{l|c|X<{\centering}|X<{\centering}}
		\toprule
		\multirow{2}{*}{Methods (Yr)} & \multirow{2}{*}{Backbone} & \multicolumn{2}{c}{Cls. Acc} \\ \cline{3-4} 
	     &  &Top-1&Top-5\\
	    \hline
	    CAM \small('16) & VGG16 & 68.8 & 88.6 \\
	    ACoL \small('18) & VGG16 & 67.5 & 88.0 \\
	    I$^2$C \small('20) & VGG16 & 69.4 & 89.3 \\
	    MEIL \small('20) & VGG16 & 70.3 & - \\
	    RCAM \small('20) & VGG16 & 68.7 & - \\
	    SLT-Net \small('21) & VGG16 & 72.4 & - \\
		\hline
	    CAM \small('16) & InceptionV3 & 73.3 & 91.8 \\
	    SPG \small('18) & InceptionV3 & 69.7 & 90.1 \\
	    ADL \small('19) & InceptionV3 & 72.8 & - \\
	    ACoL \small('18) & GoogLeNet & 71.0 & 88.2 \\
	    DANet \small('19) & GoogLeNet & 63.5 & 91.4 \\
	    RCAM \small('20) & GoogLeNet & 74.3 & - \\
	    MEIL \small('20) & InceptionV3 & 73.3 & - \\
	    I$^2$C \small('20) & InceptionV3 & 73.3 & 91.6 \\
	    SLT-Net \small('21) & InceptionV3 & \textcolor{r}{\textbf{78.1}} & - \\
	    \hline
	    TS-CAM \small('21)& Deit-S & 74.3  &  \textcolor{g}{\textbf{92.1}} \\
	    LCTR (Ours) & Deit-S & \textcolor{g}{\textbf{77.1}} & \textcolor{r}{\textbf{93.4}} \\
	    \bottomrule
    \end{tabularx}
 	\caption{Classification accuracy on the ILSVRC validation set. The results of the first and second are shown in \textcolor{r}{red} and \textcolor{g}{green}, respectively.}
  \label{tab4}
\end{table}

%----------
\textbf{Visualization.}
For qualitative evaluation, Figure~\ref{fig4} visualizes the final localization results of CAM~\cite{Zhou_2016_CVPR} based on the CNNs, TS-CAM~\cite{gao2021ts} based on the transformer and our method on CUB200-2011 and ILSVRC datasets.
From the results, compared with the CAM, we consistently observe that our method can cover a more complete range of object regions instead of focusing only on the most discriminative ones.
In addition, we capture more localized cues than the TS-CAM method, resulting in more accurate localization.
For example, the tail regions of the \textit{Bank Swallow} and the \textit{Colobus} are ignored by CAM and TS-CAM methods, while our LCTR is able to aggregate more detailed features of the target object, which enhances the local perception capability of global features among long-range feature dependencies.
Please refer to the supplementary materials for more visualized localization results of our method.

\begin{figure}[!t]
\centering
\includegraphics[width=1\linewidth]{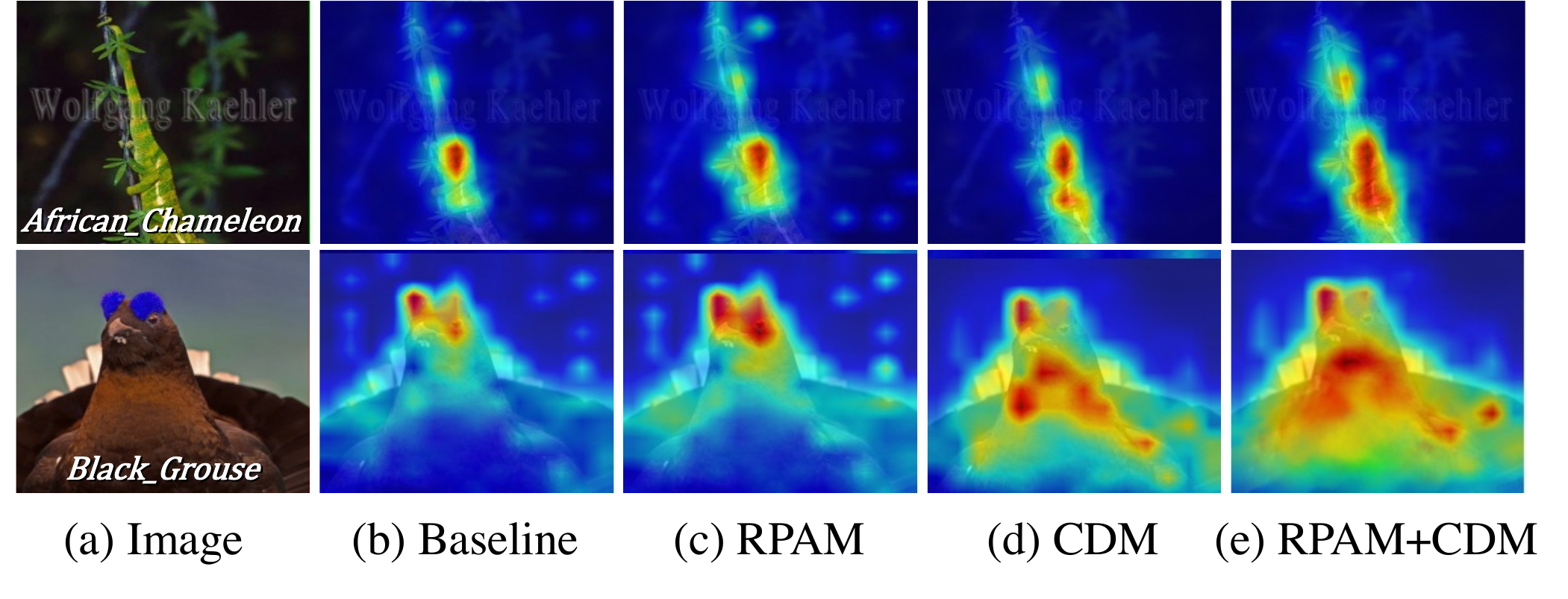}
\vspace{-20pt}
\caption{
    Visualization of localization map with different settings.
    (a) Input images.
    (b) The baseline obtains coarse long-range dependencies.
    (c) The global representations are facilitated when applying the RPAM.
    (d) The local cues are rewarded with the CDM.
    (e) The global perception capability of the target object is fully exploited.
}
\label{fig5}
\end{figure}

%%-----------ablation module.
\begin{table}[!t]
	  \centering
	    \begin{tabularx}{\linewidth}{l|ccc}
			\toprule
			Applied Mode & Top-1 \textit{Loc.} & \textit{Gt-k.} & Top-1 \textit{Cls.}\\ 
		    \hline
		    GMP-\textit{fc}  & 75.6 & 88.7 & 84.7 \\ 
			GAP-\textit{fc} & 75.8 & 88.8 & 84.8 \\
			(GMP+GAP)-\textit{fc} & 75.8 & 89.2 & 84.9 \\
			(GMP-\textit{fc}) + (GAP-\textit{fc}) & \textbf{76.0} & \textbf{90.0}& \textbf{85.0} \\
		    \bottomrule
	    \end{tabularx}
	 	\caption{The effect of different type of classifier in the CMD on CUB-200-2011 test set. GMP/GAP denotes the global max/average pooling. \textit{fc} is the fully connected layer. }
	  \label{tab5}
\end{table}

%%-----------ablation module.
\begin{table}[!t]
	  \centering
	    \begin{tabularx}{\linewidth}{X<{\centering}|X<{\centering}X<{\centering}X<{\centering}}
			\toprule
			$G$ & Top-1 \textit{Loc.} & \textit{Gt-k.} & Top-1 \textit{Cls.}\\ 
		    \hline
		    2  & 75.2 & 88.7 & 84.7 \\ 
			4 & \textbf{76.0} & \textbf{90.0} & \textbf{85.0} \\
			8 & 74.3 & 87.8 & 84.4 \\
			16 & 74.1 & 88.6 & 83.2 \\
			32 & 74.0 & 88.4 & 82.9 \\
		    \bottomrule
	    \end{tabularx}
	 	\caption{The impact of the parameter $G$ in the CDM on CUB-200-2011 test set. }
	  \label{tab6}
\end{table}

%%-----------ablation module.
\begin{table}[!t]
	  \centering
	    \begin{tabularx}{\linewidth}{c|ccc}
			\toprule
			Kernel Size ($k_w \times k_h$) & Top-1 \textit{Loc.} & \textit{Gt-k.} & Top-1 \textit{Cls.}\\ 
		    \hline
		    $1 \times 1$  & 73.5 & 88.8 & 82.5 \\ 
			$3 \times 3$  & \textbf{76.0} & \textbf{90.0} & \textbf{85.0} \\
		    \bottomrule
	    \end{tabularx}
	 	\caption{The impact of different kernel size in the CDM on CUB-200-2011 test set. }
	  \label{tab7}
\end{table}

\subsection{Ablation Studies}
First, we visualize the localization maps with different settings in Figure~\ref{fig5}.
We observe that the RPAM strengthens the global representations of the baseline~\cite{gao2021ts}, \eg, the tail-feature response of the \textit{African chameleon} is enhanced, as it considers more cross-patch information.
When only using CDM, we find that the local feature details are mined.
For example, the abdominal features of the \textit{Black grouse} are further activated compared to the baseline.
By applying both RPAM and CDM, the final localization map (Figure~\ref{fig5} (e)) highlights the full object extent.

%%-----------ablation module.
\begin{table}[!t]
	  \centering
	    \begin{tabularx}{\linewidth}{lccc|cc}
			\toprule
			\multirow{2}*{Methods} & \multirow{2}*{Dataset} & \multirow{2}*{RPAM} & \multirow{2}*{CDM} & Top-1 & Top-1 \\ 
			~ & ~ & ~ & ~ & \textit{Loc.} & \textit{Cls.}\\ 
		    \hline
		    TS-CAM  & \multirow{2}*{CUB} & & & 71.3 & 80.3 \\ 
			TS-CAM* & ~ & & & 73.1 & 81.6 \\
			\hline
			\multirow{3}*{LCTR} & \multirow{3}*{CUB} & \checkmark & & 74.0 & 81.6 \\
					& ~ & & \checkmark & 76.0 & \textbf{85.0} \\
					& ~ & \checkmark & \checkmark & \textbf{79.2} & \textbf{85.0} \\
			\hline \hline
			
			TS-CAM  & \multirow{2}*{ILSVRC} & & & 53.4 & 74.3 \\ 
			TS-CAM* & ~ & & & 53.0 & 74.0 \\
			\hline
			\multirow{3}*{LCTR} & \multirow{3}*{ILSVRC} & \checkmark & & 54.2 & 74.0 \\
					& ~ & & \checkmark & 55.1 & \textbf{77.1} \\
					& ~ & \checkmark & \checkmark & \textbf{56.1} & \textbf{77.1} \\
		    \bottomrule
	    \end{tabularx}
	 	\caption{Performance on both CUB-200-2011 test set and ILSVRC validation set when using different configurations. Note that * indicates the re-implement method.}
	  \label{tab8}
\end{table}

%----------
Next, we explore the concise design of the CDM.
From the results on Table~\ref{tab5}, we can observe that the mode of using separate \textit{fc}s with GAP and GMP reports the best performance.
These results also verify that GAP and GMP work differently in the CDM.
Then, we evaluate the accuracy under different parameters $G$ in the CDM, as shown in Table~\ref{tab6}.
From the experimental results, we observe that the best performance is achieved when $G=4$.
Setting a larger $G$ leads to a larger number of parameters and degrades accuracy, which we believe is caused by overfitting.
Besides, we examine the impact of the weighted kernel size (\ie, $k_w \times k_h$).
Results shown in Table~\ref{tab7} indicate that a kernel size of $3 \times 3$ yields better performance.

%----------
Lastly, we investigate the effect with different configurations on the accuracy, as reported in Table~\ref{tab8}.
On the CUB-200-2011 test set, we can see that RPAM increases the Top-1 \textit{Loc.} by 0.9\% compared with the baseline TS-CAM method.
Note that the lightweight RPAM is directly applied in the test phase, so the classification performance remains unchanged.
When applying the CDM to the network, we observe an improvement in both classification and localization performance.
From this, we believe that the local cues captured by the CDM are important for both two tasks.
The best localization/classification accuracy can be achieved when employing both RPAM and CDM.
Meanwhile, we conduct the similar experiments on the ILSVRC validation set, which also validate the effectiveness of two modules, as shown in the lower part of Table~\ref{tab8}.

\section{Conclusion}
In this paper, we propose a novel Local Continuity TRansformer, termed LCTR, for weakly supervised object localization, which induces the model to learn the entire extent of the object with more local cues.
We first design a relational patch-attention module (RPAM), considering cross-patch information based on the multi-head self-attention mechanism, which gathers more local patch features for facilitating the global representations.
Moreover, we introduce a cue digging module (CDM), which employs a hide-and-seek manner to wake up the weak local features for enhancing global representation learning.
Extensive experiments show the LCTR can successfully mine integral object regions and outperform the state-of-the-art localization methods.

\section{Acknowledgments}

This work is supported by the National Science Fund for Distinguished Young Scholars (No.62025603), the National Natural Science Foundation of China (No.U1705262, No.62072386, No.62072387, No.62072389, No.62002305, No.61772443, No.61802324 and No.61702136), Guangdong Basic and Applied Basic Research Foundation (No.2019B1515120049), the Natural Science Foundation of Fujian Province of China (No.2021J01002), and the Fundamental Research Funds for the central universities (No.20720200077, No.20720200090 and No.20720200091).

\bibliography{egpaper_for_review}

\end{document}